\documentclass[10pt,twocolumn,letterpaper]{article}
\pdfoutput=1
\usepackage[pagenumbers]{cvpr} 

\usepackage{graphicx}
\usepackage{amsmath}
\usepackage{amssymb}
\usepackage{booktabs}
\usepackage{adjustbox}
\usepackage{textcomp, gensymb}
\usepackage[utf8]{inputenc}
\usepackage{lipsum}
\usepackage{caption}
\usepackage{wrapfig}

\usepackage{float}
\usepackage{epsfig}
\usepackage[symbol]{footmisc}
\usepackage[pagebackref,breaklinks,colorlinks]{hyperref}

\usepackage[capitalize]{cleveref}
\crefname{section}{Sec.}{Secs.}
\Crefname{section}{Section}{Sections}
\Crefname{table}{Table}{Tables}
\crefname{table}{Tab.}{Tabs.}

\def\cvprPaperID{*****} 
\def\confName{CVPR}
\def\confYear{2023}

\begin{document}

\title{CVPR MultiEarth 2023 Deforestation Estimation Challenge: SpaceVision4Amazon}

\author{Sunita Arya\thanks{Corresponding author.} ,  S Manthira Moorthi, Debajyoti Dhar\\
Signal and Image Processing Area\\
Space Applications Centre, Ahmedabad\\
{\tt\small sunitaarya3393@gmail.com,{\tt\small \{smmoorthi,deb\}@sac.isro.gov.in}}
}
\maketitle
\begin{abstract}
In this paper, we present a deforestation estimation method based on attention guided UNet architecture using Electro-Optical (EO) and Synthetic Aperture Radar (SAR) satellite imagery. For optical images, Landsat-8 and for SAR imagery, Sentinel-1 data have been used to train and validate the proposed model. Due to the unavailability of temporally and spatially collocated data, individual model has been trained for each sensor. During training time Landsat-8 model achieved training and validation pixel accuracy of 93.45\% and Sentinel-2 model achieved 83.87\% pixel accuracy. During the test set evaluation, the model achieved pixel accuracy of 84.70\% with F1-Score of 0.79 and IoU of 0.69. 
\end{abstract}
\section{Introduction}
\label{sec:intro}
Estimation of deforestation level for Amazon Rainforest is very important for monitoring the forest change and for climate change. Degradation in forest area of Amazon can effect the global climate as Amazon Rainforest represents 40\%\cite{hubbell2008many} of tropical forest on Earth. Observation of Earth surface in any weather and lighting conditions using Electro-Optical (EO) and Synthetic Aperture Radar (SAR) sensors can be useful to monitor and analyse the change in Amazon Rainforest.
\\
Deep learning based architectures have shown impressive results in deforestation estimation task using optical as well as SAR imagery. Various models based on standard UNet and attention guided UNet architecture have been explored to segment the forest and deforested area. UNet based model has been explored in  \cite{bragagnolo2021amazon} and used it for semantic segmentation of amazon forest cover and authors of \cite{isaienkov2020deep} has also used this model for other forest ecosystem  using Sentinel-2 imagery. Landsat-8 based satellite imagery has been also used to detect the deforestation in Amazon \cite{de2020change}. \cite{torres2021deforestation} have used both Sentinel-2 and Landsat-8 images for deforestation detection in the amazon forest using fully convolutional based network. For SAR data, Sentinel-1 bands has been also used for semantic segmentation \cite{vscepanovic2021wide}.
With the advantage of attention mechanism for various computer vision tasks, authors of \cite{john2022attention} have implemented and analyzed the performance of attention UNet for semantic segmentation using Sentinel-2 satellite imagery. 
\\
For this challenge, we have used Sentinel-1 and Landsat-8 imagery provided as a part of MultiEarth 2023 Deforestation Estimation Challenge \cite{cha2023multiearth} for an attention guided UNet model. The proposed model has been tested on the given test set for evaluation.
\\
The remainder of this work is organized as follows. Section~\ref{sec:data} described about the dataset used and Section~\ref{sec:method} introduced the methodology adopted with data pre-processing and post-processing steps. The results are presented in Section~\ref{sec:result} and conclusion is shown in Section~\ref{sec:conclusion}.
\section{Dataset}
\label{sec:data}
Table~\ref{table:1} represents the description about the given training dataset for the challenge \cite{cha2023multiearth}. The spatial coverage of Amazon rain forest for this challenge is [-3.33\degree to -4.39\degree] for latitude and [-54.48\degree to -55.2\degree] for longitude. The challenge dataset consists of Sentinel-1, Sentinel-2, Landsat-5, Landsat-8 and the labels for training as shown in Table~\ref{table:1}. As the deforestation labels contains data from 2016 to 2021, we took the satellite data which are common to this range. Landsat-5 data has been discarded for this work because of not common range of year. Out of Landsat-8 and Sentinel-2 optical images, we took only Landsat-8 data. Sentinel-1 data to handle the cloudy images. Sentinel-1 has two polarization bands: VV and VH with spatial resolution of 10m and Landsat-8 has seven surface reflectance bands with one surface temperature band at a spatial resolution of 30m. 
\begin{table*}[!htb]
\centering
\begin{tabular}{c c c c c} 
 \hline
 \textbf{Satellite} & \textbf{Temporal Coverage} & \textbf{Spatial Resolution} & \textbf{Band} & \textbf{Image Size} \\ [0.5ex] 
 \hline 
\textbf{Sentinel-1} & 2014-2021 & 10m & VV,VH & 256x256 \\ 
 \textbf{Sentinel-2} & 2018-2021 & 10m & B1-B12, QA60  & 256x256 \\
 \textbf{Landsat-5} & 1984-2012 & 30m  & SR\_B1-SR\_B7, QA\_PIXEL & 85x85 \\
 \textbf{Landsat-8} & 2013-2021 & 30m  & SR\_B1-SR\_B7, ST\_B10, QA\_PIXEL & 85x85 \\
 \textbf{Labels} & 2016-2021 & 10m & Binary Mask Band [0,1] & 256x256 \\[1ex] 
 \hline
\end{tabular}
\caption{Dataset for challenge.}
\label{table:1}
\end{table*}
\begin{table*}[!htb]
\centering
\begin{tabular}{c c c c c c} 
 \hline
 \textbf{Satellite} & \textbf{Spatial Resolution} & \textbf{Band} & \textbf{Image Size} & \textbf{Total Samples}\\ [0.5ex] 
 \hline 
\textbf{Sentinel-1} & 10m & VV, VH, VV/VH & 256x256x3 & 18014\\ 
 \textbf{Landsat-8} &  10m  & SR\_B1-SR\_B7, ST\_B10 & 256x256x7 & 6313\\
 \textbf{Labels} &  10m & Binary Mask Band [0,1] & 256x256 & Paired \\[1ex] 
 \hline
\end{tabular}
\caption{Final Dataset for training the model.}
\label{table:2}
\end{table*}
\begin{figure*}[!htb]
    \centering
    \includegraphics[width=0.75\textwidth]{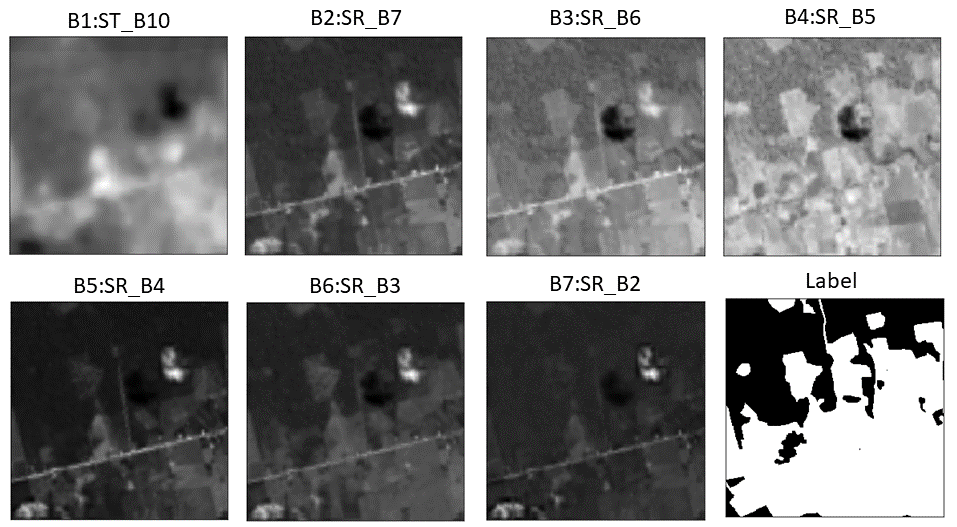}
    \caption{Landsat-8 data for training.}
    \label{landsat8_input}
\end{figure*}
\begin{figure*}[!htb]
    \centering
    \includegraphics[width=0.75\textwidth]{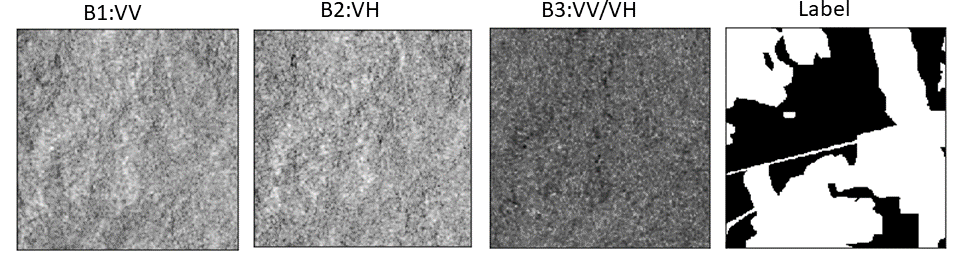}
    \caption{Sentinel-1 data for training.}
    \label{sentinel1_input}
\end{figure*}
\\
\\
\\
After careful analysis of the provided challenge data, we finally considered Landsat-8 for optical imagery and Sentinel-1 for SAR imagery at a unified resolution. Table~\ref{table:2} represents the description about the dataset taken for training the model.
\section{Methodology}
\label{sec:method}
\begin{figure*}[ht]
    \centering
    \includegraphics[width=0.9\textwidth]{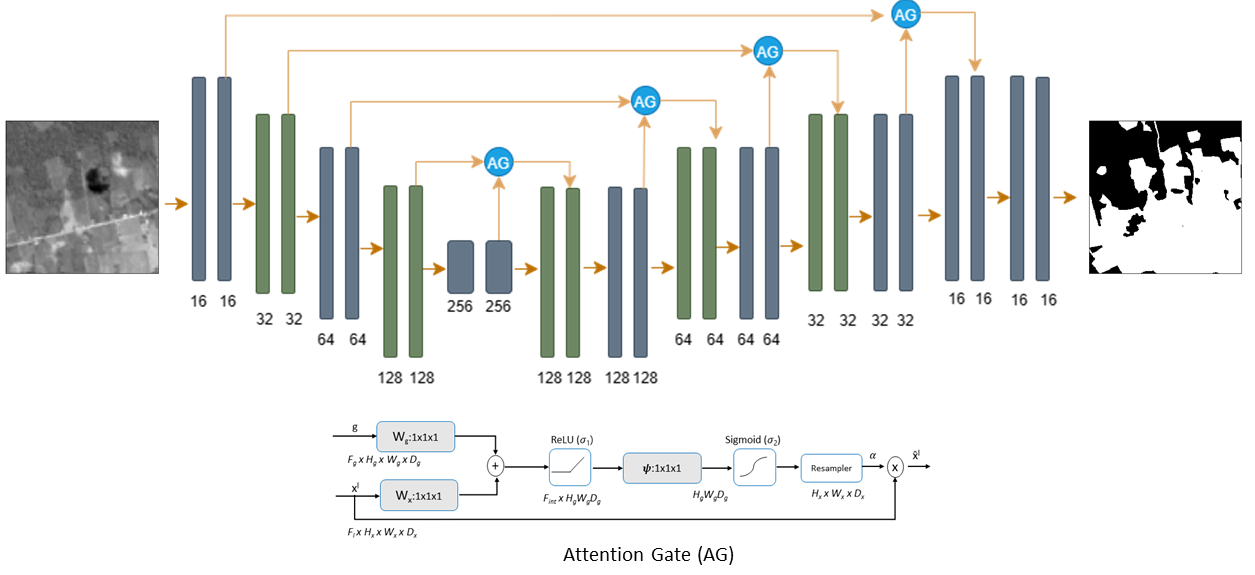}
    \caption{Attention UNet model with attention gate same as in \cite{oktay2018attention}}
    \label{model_1}
\end{figure*}
This sections describes about the data pre-processing steps, model architecture with training details and finally data post-processing steps taken to refine the generated results.
\subsection{Data Pre-Processing}
\textbf{Landsat-8:} In electromagnetic spectrum range, every spectral channel has it's own significance for specific target. So, rather than considering only major vegetation influencing bands, we took all the spectral bands for this work. We considered both surface reflectance and surface temperature bands. For training, all the bands have been normalized between 0 and 1. Additionally, we resampled the data to make unified resolution with the spatial resolution of given label images. After data pre-processing step, total number of training samples taken are 6313.  \\
\textbf{Sentinel-1:} For Sentinel-1, we considered both the given bands i.e. VV and VH bands to train and validate the model. In addition to that we have taken a third band by taking ratio of both the given channels. We have used 1\% percentage bandwise contrast stretching as suggested in \cite{vscepanovic2021wide} for normalizing the SAR data and then converting the it to [0-1] for training. Samples taken for final training the model are 18014. 
\subsection{Network and Training Details}
To segment the forest and deforested area in this work, we took pairwise training images with their respective labels as described in Section 2. Our model is inspired by the \cite{john2022attention} work. Figure~\ref{model_1} represents the model parameters including attention gate. For training, we used batch size of 16, Adam optimizer \cite{kingma2014adam} with learning rate of 0.0001 for 50 epochs. To handle the class imbalance, we combined Binary Cross-Entropy loss \cite{yi2004automated} with Dice loss \cite{sudre2017generalised} with an equal weight to both the losses. Additionally, data augmentation such as rotation, horizontal and vertical flip have been used for Landsat-8 model only.
\subsection{Output Refinement}
\begin{enumerate}
    \item \textbf{Indices Based }: For Landsat-8 generated results, we have used Normalized Difference Vegetation Index (NDVI) for discarding the cloudy images mask. We took 0.1 as a threshold value to generate the NDVI based cloud mask. For test images which has value greater than 1\%, we discarded it as cloudy image and did not consider it for the next step of deforestation mask generation.
    \item \textbf{Morphological Operator}: After discarding the cloudy images for Landsat-8,  we averaged out all the remaining masks taking 0.4 as a threshold for deforestation mask generation. Finally, we used morphological operator erosion followed by dilation to refine the final masks as explored in \cite{lee2022multiearth}. For Sentinel-1, we averaged the generated masks for given test query and applied the morphological operators in same order. 
\end{enumerate}
\section{Results}
\label{sec:result}
\begin{figure*}[!htb]
    \centering
    \includegraphics[width=0.86\textwidth]{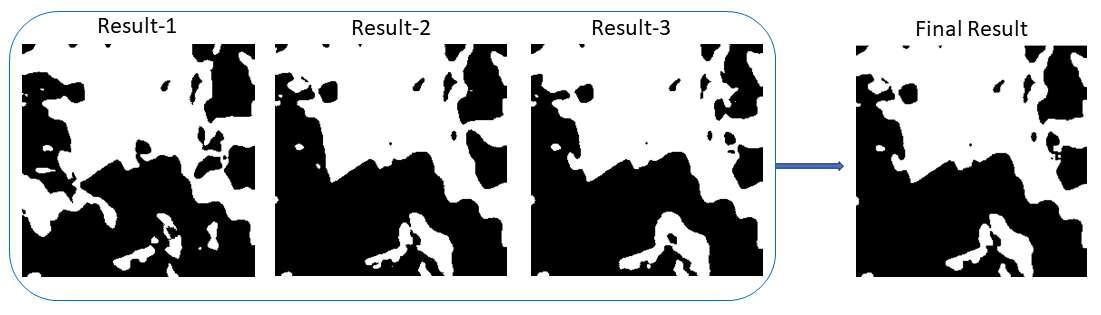}
    \caption{Landsat-8 results on test set.}
    \label{landsat8_result}
\end{figure*}

\begin{figure*}[!htb]
    \centering
    \includegraphics[width=0.86\textwidth]{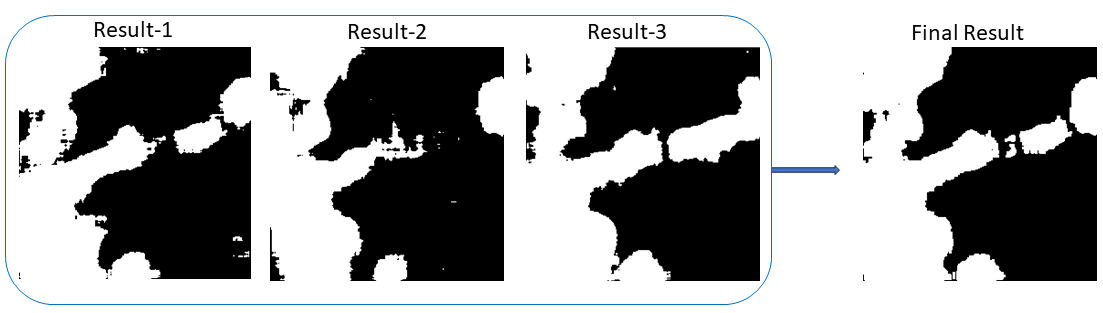}
    \caption{Sentinel-1 results on test set.}
    \label{sentinel1_result}
\end{figure*}

For evaluation, additional data of both the sensors has given with the test queries. The test set consists of 1000 queries from August 2016 to August 2021 for latitude range from -3.87 to -4.39 and longitude range from -54.8 to -54.88. There are the cases where few locations and dates were not available for Landsat-8, but available in Sentinel-1 data. Table~\ref{table:3} represents the results of test queries on the evaluation server. Figure~\ref{landsat8_result} represents final results using Landsat-8 model and Figure~\ref{sentinel1_result} shows results of Sentinel-1 model.\\
\begin{table*}[!htb]
\centering
\begin{tabular}{c c c c } 
 \hline
 \textbf{Submission Version} & \textbf{Pixel Accuracy} & \textbf{F1-Score} & \textbf{IoU}  \\ [0.5ex] 
 \hline 
\textbf{SpaceVision4Amazon} & 83.14 & 0.7925 & 0.691 \\ 
\textbf{SpaceVision4Amazon\_v2} & 84.70 & 0.7950 & 0.694 \\[1ex] 
 \hline
\end{tabular}
\caption{Final performance on test dataset.}
\label{table:3}
\end{table*}
\\
\textbf{\textit{SpaceVision4Amazon}}: For submission version \textit{SpaceVision4Amazon} we directly averaged out all results of available images for the given test queries using Landsat-8 and Sentinel-1 imagery.
\\
\textbf{\textit{SpaceVision4Amazon\_v2}}: For submission version \textit{SpaceVision4Amazon\_v2} we followed the output refinement steps as discussed in Section 3 for both Landsat-8 and Sentinel-1 dataset.\\
Model is implemented in Python v3.7.4 using Keras \cite{chollet2015keras} v2.4.3 with Tensorflow \cite{abadi2016tensorflow} v2.3.1 backend, and hardware configuration with CPU Intel(R) Xeon(R) Platinum 8180 CPU @ 2.50GHz with 1TB of memory and Quadro V100 GPU with 32GB memory.

\section{Conclusion}
\label{sec:conclusion}
In this paper, we present a deforestation estimation method for Amazon rainforest under the MultiEarth 2023 sub challenge. Our method based on attention guided UNet architecture for provided Optical and Synthetic Aperture Radar (SAR) satellite imagery. For optical images, Landsat-8 and for SAR imagery, Sentinel-1 data have been used to train and validate the proposed model. The model achieved pixel accuracy of 84.70\% with F1-Score of 0.79 and IoU of 0.69 on the test set for evaluation. 
{\small
\bibliographystyle{ieee_fullname}

\begin{thebibliography}{10}\itemsep=-1pt

\bibitem{abadi2016tensorflow}
Mart{\'\i}n Abadi, Ashish Agarwal, Paul Barham, Eugene Brevdo, Zhifeng Chen,
  Craig Citro, Greg~S Corrado, Andy Davis, Jeffrey Dean, Matthieu Devin, et~al.
\newblock Tensorflow: Large-scale machine learning on heterogeneous distributed
  systems.
\newblock {\em arXiv preprint arXiv:1603.04467}, 2016.

\bibitem{bragagnolo2021amazon}
Ld Bragagnolo, Roberto~Valmir da Silva, and Jos{\'e} Mario~Vicensi Grzybowski.
\newblock Amazon forest cover change mapping based on semantic segmentation by
  u-nets.
\newblock {\em Ecological Informatics}, 62:101279, 2021.

\bibitem{cha2023multiearth}
Miriam Cha, Gregory Angelides, Mark Hamilton, Andy Soszynski, Brandon Swenson,
  Nathaniel Maidel, Phillip Isola, Taylor Perron, and Bill Freeman.
\newblock Multiearth 2023 -- multimodal learning for earth and environment
  workshop and challenge, 2023.

\bibitem{chollet2015keras}
Fran\c{c}ois Chollet et~al.
\newblock Keras.
\newblock \url{https://keras.io}, 2015.

\bibitem{de2020change}
Pablo~Pozzobon De~Bem, Osmar~Ab{\'\i}lio de Carvalho~Junior, Renato
  Fontes~Guimar{\~a}es, and Roberto~Arnaldo Trancoso~Gomes.
\newblock Change detection of deforestation in the brazilian amazon using
  landsat data and convolutional neural networks.
\newblock {\em Remote Sensing}, 12(6):901, 2020.

\bibitem{hubbell2008many}
Stephen~P Hubbell, Fangliang He, Richard Condit, Lu{\'\i}s Borda-de {\'A}gua,
  James Kellner, and Hans Ter~Steege.
\newblock How many tree species are there in the amazon and how many of them
  will go extinct?
\newblock {\em Proceedings of the National Academy of Sciences},
  105(supplement\_1):11498--11504, 2008.

\bibitem{isaienkov2020deep}
Kostiantyn Isaienkov, Mykhailo Yushchuk, Vladyslav Khramtsov, and Oleg
  Seliverstov.
\newblock Deep learning for regular change detection in ukrainian forest
  ecosystem with sentinel-2.
\newblock {\em IEEE Journal of Selected Topics in Applied Earth Observations
  and Remote Sensing}, 14:364--376, 2020.

\bibitem{john2022attention}
David John and Ce Zhang.
\newblock An attention-based u-net for detecting deforestation within satellite
  sensor imagery.
\newblock {\em International Journal of Applied Earth Observation and
  Geoinformation}, 107:102685, 2022.

\bibitem{kingma2014adam}
Diederik~P Kingma and Jimmy Ba.
\newblock Adam: A method for stochastic optimization.
\newblock {\em arXiv preprint arXiv:1412.6980}, 2014.

\bibitem{lee2022multiearth}
Dongoo Lee and Yeonju Choi.
\newblock Multiearth 2022 deforestation challenge--forestgump.
\newblock {\em arXiv preprint arXiv:2206.10831}, 2022.

\bibitem{oktay2018attention}
Ozan Oktay, Jo Schlemper, Loic~Le Folgoc, Matthew Lee, Mattias Heinrich,
  Kazunari Misawa, Kensaku Mori, Steven McDonagh, Nils~Y Hammerla, Bernhard
  Kainz, et~al.
\newblock Attention u-net: Learning where to look for the pancreas.
\newblock {\em arXiv preprint arXiv:1804.03999}, 2018.

\bibitem{vscepanovic2021wide}
Sanja {\v{S}}{\'c}epanovi{\'c}, Oleg Antropov, Pekka Laurila, Yrjo Rauste,
  Vladimir Ignatenko, and Jaan Praks.
\newblock Wide-area land cover mapping with sentinel-1 imagery using deep
  learning semantic segmentation models.
\newblock {\em IEEE Journal of Selected Topics in Applied Earth Observations
  and Remote Sensing}, 14:10357--10374, 2021.

\bibitem{sudre2017generalised}
Carole~H Sudre, Wenqi Li, Tom Vercauteren, Sebastien Ourselin, and M
  Jorge~Cardoso.
\newblock Generalised dice overlap as a deep learning loss function for highly
  unbalanced segmentations.
\newblock In {\em Deep Learning in Medical Image Analysis and Multimodal
  Learning for Clinical Decision Support: Third International Workshop, DLMIA
  2017, and 7th International Workshop, ML-CDS 2017, Held in Conjunction with
  MICCAI 2017, Qu{\'e}bec City, QC, Canada, September 14, Proceedings 3}, pages
  240--248. Springer, 2017.

\bibitem{torres2021deforestation}
Daliana~Lobo Torres, Javier~Noa Turnes, Pedro~Juan Soto~Vega, Raul~Queiroz
  Feitosa, Daniel~E Silva, Jose Marcato~Junior, and Claudio Almeida.
\newblock Deforestation detection with fully convolutional networks in the
  amazon forest from landsat-8 and sentinel-2 images.
\newblock {\em Remote Sensing}, 13(24):5084, 2021.

\bibitem{yi2004automated}
Ma Yi-de, Liu Qing, and Qian Zhi-Bai.
\newblock Automated image segmentation using improved pcnn model based on
  cross-entropy.
\newblock In {\em Proceedings of 2004 International Symposium on Intelligent
  Multimedia, Video and Speech Processing, 2004.}, pages 743--746. IEEE, 2004.

\end{thebibliography}

}

\end{document}